\documentclass[10pt]{article}

\usepackage{xcolor}
\usepackage{amsmath}
\usepackage{float}
\usepackage{graphicx}
\usepackage{tabularx}
\usepackage[letterpaper, top = 1.0in, bottom = 1.0in, right=0.75in, left = 0.75in]{geometry}
\usepackage{amssymb} 
\usepackage{titling}
\usepackage{titlesec}
\usepackage{times}
\usepackage{ragged2e}

\usepackage{blindtext}
\usepackage{graphics} 
\usepackage{epsfig} 
\usepackage{bm}
\usepackage{caption}
\usepackage{subcaption}
\usepackage[noadjust]{cite}
\usepackage[labelformat=simple]{subcaption}

\usepackage{dsfont}
\usepackage{relsize}
\usepackage{verbatim}

\DeclareMathOperator*{\argmin}{argmin}

\pretitle{ \begin{center} \Huge}

\posttitle{\end{center}}
\preauthor{\begin {center} \normalsize}

\postauthor{\end{center}}
\date{}
\titleformat{\section}{\normalsize \bfseries \scshape}{\thesection}{1em}{}
\titleformat{\subsection}{\normalsize \bfseries}{\thesubsection}{1em}{}
\title{\Huge SLAM-based Integrity Monitoring \\ Using GPS and Fish-eye Camera}

\author{Sriramya Bhamidipati,~\textit{University of Illinois at Urbana-Champaign} \\
	Grace Xingxin Gao,~\textit{Stanford University} }

\begin{document}
\maketitle
\thispagestyle{empty}

\section*{Biographies}
\noindent \textbf{Sriramya Bhamidipati} is a Ph.D. student in the Department of Aerospace Engineering at the University of Illinois at Urbana-Champaign, where she also received her master’s degree in 2017. She obtained her B.Tech. in Aerospace from the Indian Institute of Technology Bombay in 2015. Her research interests include GPS, power and control systems, artificial intelligence, computer vision and unmanned aerial vehicles. 
~\\

\noindent \textbf{Grace Xingxin Gao} is an assistant professor in the Department of Aeronautics and Astronautics at Stanford University. Before joining Stanford University, she was an assistant professor at University of Illinois at Urbana-Champaign. She obtained her Ph.D. degree at Stanford University. Her research is on robust and secure positioning, navigation and timing with applications to manned and unmanned aerial vehicles, robotics and power systems.

\section*{Abstract}
Urban navigation using GPS and fish-eye camera suffers from multipath effects in GPS measurements and data association errors in pixel intensities across image frames. 
We propose a Simultaneous Localization and Mapping (SLAM)-based Integrity Monitoring (IM) algorithm to compute the position protection levels while accounting for multiple faults in both GPS and vision. 
We perform graph optimization using the sequential data of GPS pseudoranges, pixel intensities, vehicle dynamics and satellite ephemeris to simultaneously localize the vehicle as well as the landmarks, namely GPS satellites and key image pixels in the world frame. We estimate the fault mode vector by analyzing the temporal correlation across the GPS measurement residuals and spatial correlation across the vision intensity residuals. In particular, to detect and isolate the vision faults, we developed a superpixel-based piecewise Random Sample Consensus~(RANSAC) technique to perform spatial voting across image pixels. For an estimated fault mode, we compute the protection levels by applying worst-case failure slope analysis to the linearized Graph-SLAM framework.

We perform ground vehicle experiments in the semi-urban area of Champaign, IL and have demonstrated the successful detection and isolation of multiple faults. We also validate tighter protection levels and lower localization errors achieved via the proposed algorithm as compared to SLAM-based IM that utilizes only GPS measurements. 

\section{Introduction} \label{sec:intro}
Integrity Monitoring~(IM) serves as a important performance metric to assess the navigation solution estimation~\cite{ochieng2003gps}. Vehicles operating in urban areas face challenges~\cite{joerger2017towards} due to static infrastructure, such as buildings and thick foliage, dynamic obstacles, such as traffic and pedestrians, and environmental conditions, such as shadows, sunlight and weather. GPS systems receive fewer measurements in urban environments due to degraded satellite visibility. They also suffer from received signal faults caused by multipath and satellite faults caused by anomalies in the broadcast navigation message. 

To address the above-mentioned challenges, one possible solution is to incorporate additional redundancy through the sensor fusion of GPS and vision. Vision sensor performs well in urban areas due to the feature-rich surroundings~\cite{hol2011sensor}. Sensor fusion~\cite{krishnaswamy2008sensor} integrates measurements from multiple sensors to improve the accuracy of the vehicle and provide robust performance. Individual sensors, such as GPS and camera, have inherent limitations in operability that are reliably corrected by combining these complementary sensors in a sensor fusion framework. In particular, occlusion and illumination variations in multiple pixel intensities induce data association errors across images, thereby termed as vision faults~\cite{miro2006towards}. Therefore, there is a need for the development of sensor-fusion-based IM techniques that account for multiple faults in both GPS and vision. 
 
Rich literature exists on urban IM approaches for GPS-based navigation systems that utilize external information sources. In~\cite{velaga2012map}, the authors developed a sequential map-aided IM technique that checks for outliers in position and Geographic Information System~(GIS) using traditional RAIM~\cite{walter2008shaping} and weight-based topological map-matching process, respectively. Another paper~\cite{binjammaz2013gps} developed three phases of integrity checks that include assessing the position quality via traditional Receiver Autonomous Integrity Monitoring~(RAIM), speed integrity via GPS Doppler and map matching accuracy via fuzzy inference system. However, these approaches have practical limitations because the offline map database is not always available and its accuracy cannot be guaranteed due to the dynamic changes in the urban surroundings. Another line of prior work~\cite{li2017lane,toledo2009lane} utilizes the odometry information obtained from Dead-Reckoning~(DR) sensors, such as inertial measurement units, wheel speed encoder and camera, to perform GPS-based IM. But the drawbacks of these approaches are that they do not address the faults associated with DR sensors, and also, do not account for the simultaneous occurrence of faults across multiple sensor sources.  

In this paper, we leverage the generalized and flexible platform developed in our prior work~\cite{bhamidipati2018multiple}, which is Simultaneous Localization and Mapping~(SLAM)-based Fault Detection and Isolation~(FDI), as the basis for assessing the sensor fusion integrity. Another extension of the SLAM-based FDI platform, described in our prior work~\cite{bhamidipati2018receivers}, assesses the integrity of cooperative localization using a network of receivers. SLAM~\cite{cadena2016past}, a well-known technique in robotics, utilizes sensor measurements to estimate the landmarks in a three-dimensional~(3D) map while simultaneously localizing the robot within it. Analogous to this, our prior work~\cite{bhamidipati2018multiple} on SLAM-based FDI combines the sequential data of GPS measurements, receiver motion model and satellite orbital model in a graph framework to simultaneously localize the \textit{robot}, which is the GPS receiver, \textit{landmarks} in the map, which are the GPS satellites. A key feature of this platform is that it utilizes graph optimization techniques~\cite{latif2014robust} and therefore, does not require any prior assumption regarding the distribution of states. Given that we localize the landmarks as well, our SLAM-based FDI does not require any prior information regarding the surrounding 3D maps.  

We propose SLAM-based IM algorithm using GPS and fish-eye camera to compute the error bounds, termed as \textit{protection levels}, of the estimated navigation solution by applying worst-case failure slope analysis~\cite{salos2013receiver,joerger2014solution} to the Graph-SLAM framework. In this work, we consider \textit{global landmarks} as the GPS satellites and additional \textit{local landmarks} as the key image pixels in the world frame. Here, world frame represents the Earth-Centered Earth-Fixed~(ECEF) frame. We simultaneously update the state vectors of the vehicle, GPS satellites and key image pixels and thereafter, perform multiple FDI. We constrain the graph via GPS pseudoranges, raw fish-eye images, vehicle dynamics and satellite ephemeris. 

For vision measurements, we opt for a fish-eye camera mounted on an vehicle and point it upwards for the following reasons; Firstly, given its wide~$(\geq 180^{\circ})$ Field-Of-View~(FOV) the image pixels are spatially spread-out in different directions with respect to the vehicle, thereby, compensating for the restricted spatial geometry of the limited global landmarks, i.e., GPS satellites. Secondly, given that the camera is pointing upwards, the unstructured skyline of the buildings aids in resolving the attitude of the vehicle. Thirdly, the fish-eye image captures the open sky section with respect to the vehicle that is used to distinguish the Line-Of-Sight~(LOS) GPS satellites from that of the Non-Line-Of-Sight~(NLOS) ones~\cite{shytermeja2014proposed}. 
\begin{figure}[H]
	\setlength{\belowcaptionskip}{-4pt}
	\centering	\includegraphics[width=0.6\columnwidth]{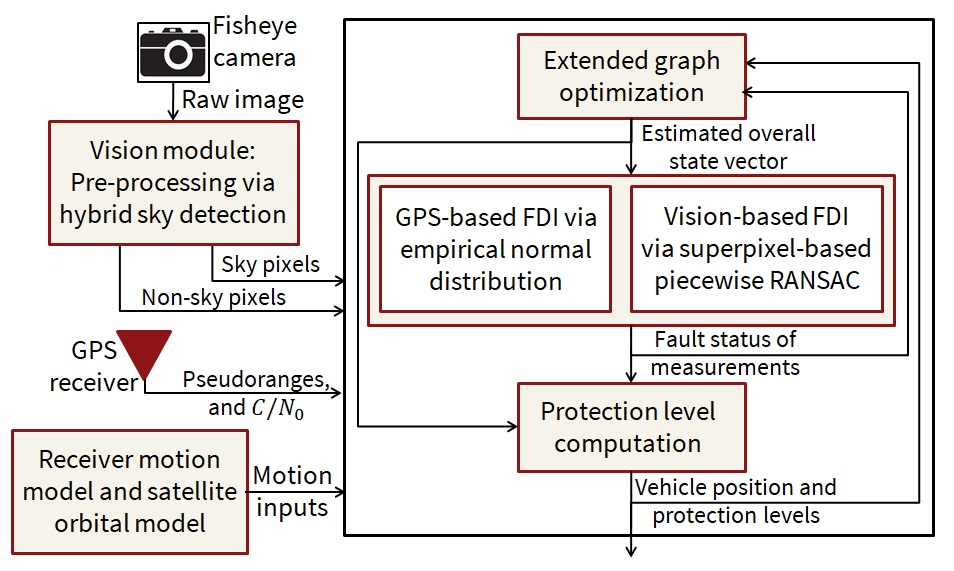}
	\caption{Architecture of our SLAM-based IM algorithm using GPS and fish-eye camera.}
	\label{fig_algo:architecture}
\end{figure}
The rest of the paper is organized as follows:~Section~II describes our SLAM-based IM algorithm that utilizes GPS and fish-eye camera; Section~III experimentally validates the proposed algorithm in performing multiple FDI of GPS and vision faults and assessing the corresponding localization accuracy and size of protection levels;~Section~IV concludes the paper.

\section{SLAM-based IM using GPS and Fish-eye Camera}
We outline the high-level architecture and later, explain the details of the proposed algorithm. 
In this work, we focus on the measurement faults that are more frequently observed in urban areas, namely GPS and vision faults. Even though the formulation of the proposed algorithm is capable of addressing other faults, for simplicity, we consider no measurement faults associated with the receiver motion model and satellite orbital model. For reference, details regarding addressing the satellite faults using SLAM-based FDI are described in our prior work~\cite{bhamidipati2018multiple}. 
In Fig.~\ref{fig_algo:architecture}, we show the architecture of our SLAM-based IM algorithm using GPS and fish-eye camera that is summarized as follows: 
\begin{enumerate}
	\item During initialization, we initialize a 3D map using the PVT of the receiver and satellites computed via an established GPS receiver algorithm~\cite{lashley2010valid}. We set the initial value of all GPS measurement fault status to $0.5$ indicating neutrality. For the vision module, we perform initial calibration to estimate the scaling from image to global frame.
	\item Firstly, we pre-process the raw image obtained from the fish-eye camera using our hybrid sky detection algorithm to distinguish the sky pixels from the non-sky pixels. The detected sky pixels are used to distinguish the LOS and NLOS satellites and thereafter, formulate the corresponding GPS measurement covariance. 
	\item We consider the non-sky pixels along with GPS pseudoranges and carrier-to-noise density~$(C/N_{0})$ values, receiver motion model and satellite orbital model as input measurements to our algorithm. We combine the measurements in an extended graph optimization module to estimate the overall state vector, which consists of the state vector of the vehicle, GPS satellites and key image pixels using M-estimator~\cite{shevlyakov2008redescending}-based Levenberg Marquardt algorithm~\cite{lourakis2005brief}. 
	\item We independently analyze the measurement residuals against an empirical distribution to detect and isolate GPS faults. We develop a superpixel~\cite{li2015superpixel}-based piecewise Random Sample Consensus~(RANSAC)~\cite{conte2009vision} to perform spatial voting for the detection and isolation of vision faults. Based on the estimated fault status of the measurements, we estimate the measurement fault mode, which has binary entries, such that $0$ indicates non-faulty and $1$ represents faulty. 
	\item Finally, utilizing the estimated fault mode and overall state vector, we formulate the failure slope for the Graph-SLAM framework and subsequently, compute the protection levels using worst-case failure mode slope analysis~\cite{salos2013receiver,joerger2014solution}. 
\end{enumerate}
\begin{figure}[H]
	\centering
	\begin{subfigure}[b]{0.75\textwidth}
		\includegraphics[width=\textwidth]{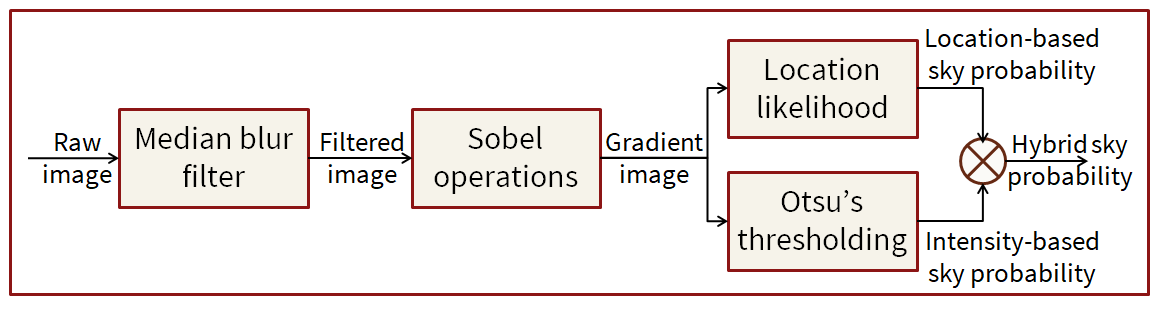}
		\caption{Our hybrid sky detection algorithm}
		\label{fig_exp:hybrid_sky}
	\end{subfigure}
	\begin{subfigure}[b]{0.255\textwidth}
		\includegraphics[width=\textwidth]{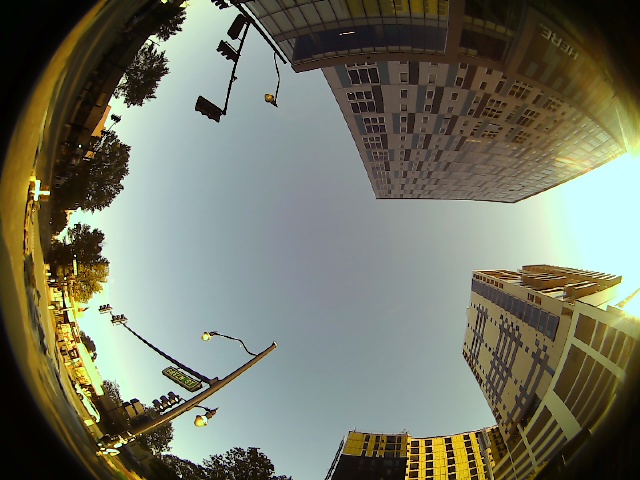}
		\caption{Raw fish-eye image}
		\label{fig_exp:raw}
	\end{subfigure}
	\begin{subfigure}[b]{0.255\textwidth}
		\includegraphics[width=\textwidth]{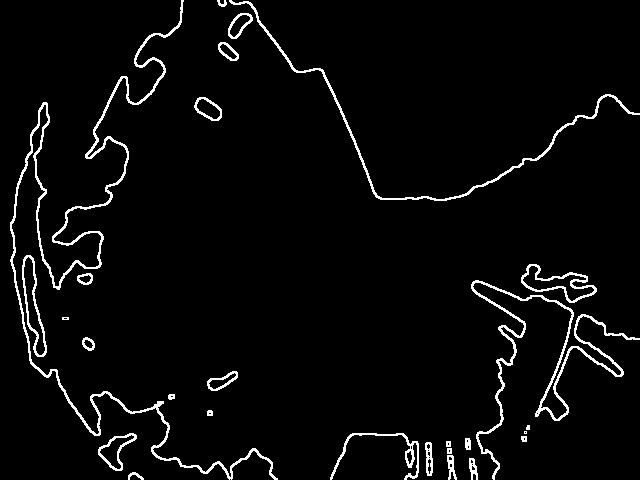}
		\caption{Hybrid optimal border}
		\label{fig_exp:border}
	\end{subfigure}
	\begin{subfigure}[b]{0.255\textwidth}
		\includegraphics[width=\textwidth]{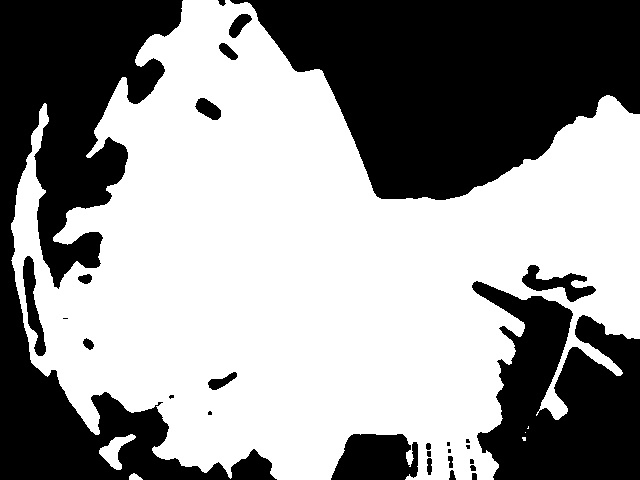}
		\caption{Sky area}
		\label{fig_exp:area}
	\end{subfigure}
	\caption{An example showing the pipeline of pre-processing the fish-eye image in the vision module}
	\label{fig_algo:vision}
\end{figure}

The proposed SLAM-based IM algorithm using GPS and fish-eye camera consists of three main modules, namely measurement pre-processing, extended graph optimization and IM for Graph-SLAM. We describe the details as follows:

\subsection{Pre-processing the measurements}
We consider the following measurements as inputs to our SLAM-based IM algorithm: GPS pseudoranges and $C/N_{0}$ values from the GPS receiver, pixel intensities from the fish-eye image, control input obtained from the vehicle motion model and satellite ephemeris decoded from the navigation message. 

~\\
\noindent \textit{1. Vision module: }~\\ 
\indent We pre-process the raw image obtained from fish-eye camera using hybrid sky detection algorithm, to distinguish the sky-pixels from the non-sky pixels. The pipeline of our hybrid sky detection is seen in Fig.~\ref{fig_exp:hybrid_sky}. Our hybrid sky detection takes into account not only the pixel intensities but also prior knowledge regarding the spatial location of the sky pixels. 

We convert the raw image to gray scale and then perform median blur operation. The median blur~\cite{wang2010new} is a non-linear filter that reduces the noise in the image while keeping the edges relatively sharp. Next, we compute the gradient by combining the magnitude obtained via two Sobel operations~\cite{gao2010improved}, one executed in horizontal and the other in vertical direction. An example of the image obtained from fish-eye camera operating in urban areas and pointing upwards is seen in Fig.~\ref{fig_exp:raw}. 

We observe that the probability of sky is highest close to the center and exponentially decreases outwards~\cite{haque2008hybrid}. Therefore, the corresponding location-based sky probability, denoted by~$p_{loc}$ is given by
\begin{align} \label{eq_algo:loc}
    p_{loc}({\bm u}) = \textrm{exp}\Big(-\dfrac{2\big|\big| {\bm u}-{\bm c}_{loc}\big|\big|}{|\Pi|}\Big),
\end{align}
\noindent where ${\bm u}$ is the 2D image coordinates, such that ${\bm u}=[u,v]^{T}\in \Pi$, $\Pi$~represents the pre-defined domain of the image coordinates. $|\cdot|$~denotes the cardinality of the image domain and $||\cdot||$~denotes the 2-norm residual. ${\bm c}_{loc}\subset \Pi$ denotes the pre-defined center coordinates in the 2D image frame.

Combining the location probability with Otsu's method of intensity thresholding~\cite{moghaddam2012adotsu}, we compute the hybrid optimal border, seen in Fig.~\ref{fig_exp:border}, that separates the sky region, represented by subscript~$sky$, from that of the infrastructure, denoted by subscript~$inf$. We minimize the variance of sky and infrastructure to estimate the Otsu's intensity threshold~$I_{otsu}$ as 
\begin{align} \label{eq_algo:otsu}
I_{otsu}= \argmin_{k\in\,\bm{I}}\Big(\omega_{sky}(k)\sigma _{sky}^{2}(k)+\omega_{inf}(k)\sigma_{inf}^{2}(k) \Big)~\textrm{from~\cite{moghaddam2012adotsu}},
\end{align}
\noindent where
\begin{itemize}
	\item [--] ${\bm I}$~denotes the intensity vector that stacks all the pixel intensities in the fish-eye image, such that ${\bm I}=\{I({\bm u})\,\big|\, {\bm u}\in \Pi \}$, where $I({\bm u}): \Pi \rightarrow \mathbb{R}$~denotes the intensity of any 2D pixel coordinates~${\bm u}$; 
    \item [--] $\omega _{sky}(k)$ and $\omega _{inf}(k)$ denotes the weights associated with the sky and building infrastructure, respectively, such that $\omega _{sky}(k)=\dfrac{\sum_{{\bm u}\in |\Pi|}\mathds{1}\big\{I({\bm u})<k\big\}}{|\Pi|}$ and $\omega _{inf}(k)=\dfrac{\sum_{{\bm u}\in |\Pi|}\mathds{1}\big\{I({\bm u})>k\big\}}{|\Pi|}$;
    \item[--] $\sigma_{sky}^{2}(k)$ and $\sigma_{inf}^{2}(k)$ denotes the variance of the pixel intensities associated with the sky and building infrastructure. 
\end{itemize}

Utilizing Eqs.~\eqref{eq_algo:loc} and~\eqref{eq_algo:otsu}, we compute the hybrid sky probability, denoted by $p_{sky}$ at any 2D image coordinate~${\bm u},\,{\bm u}\in\Pi$ as
\begin{align} \label{eq_algo:hybrid}
p_{sky}({\bm u}) &= \textrm{exp}\bigg(\dfrac{-\big| I({\bm u})-I_{otsu}\big|}{\big|I_{max}-I_{min}\big|}\bigg)~p_{loc}({\bm u}),
\end{align}
\noindent where $I_{max}$ and $I_{min}$ are the maximum and minimum intensity values in the fish-eye image, such that $I_{max}=\max_{{\bm u}\in \Pi}I({\bm u})$ and $I_{min}=\min_{{\bm u}\in \Pi}I({\bm u})$, respectively. Considering $\eta$ as the pre-defined sky threshold, if $p_{sky}({\bm u})>\eta$, then it is categorized as sky pixel and non-sky pixel otherwise. The sky-enclosed area in the fish-eye image is seen in  Fig.~\ref{fig_exp:area}.   

Next, using the non-sky detected pixels, we describe the vision measurement model in Eq.~\eqref{eq_ion:cam_model} that is formulated via omni-directional camera model~\cite{caruso2015large} and direct image alignment~\cite{engel2014lsd}. Direct image alignment computes the depth maps in an incremental fashion and compares the pixel intensities across the image frames directly, such that the spatial context of the image is preserved. This vision measurement model is utilized later in our extended graph optimization module to formulate the corresponding vision odometry-based component of the cost function. 

\begin{align}
\begin{split} \label{eq_ion:cam_model}
I_{kf}({\bm u}) &= I_{t}\Big(\pi\big(w(\Delta{\bm \mu}_{t},{\bm u})\big)\Big) + \eta_{vis}({\bm u})~\textrm{from~\cite{engel2014lsd}},
\end{split}
\end{align} 
\noindent such that $\eta_{vis}({\bm u})$ is pixel noise and from~\cite{caruso2015large}, 
\begin{align*}
\begin{split}
w(\Delta{\bm \mu},{\bm u}) &= {\textbf R}(\Delta{\bm \mu})\,\pi^{-1}\Big({\bm u},d_{kf}({\bm u})\Big) + {\textbf t}(\Delta{\bm \mu}), \\
\pi(\textbf{p}) &= 
\begin{bmatrix}
\vspace{0.5pc}
f_{x}\dfrac{p_{x}}{p_{z}+\big|\big|\textbf{p}\big|\big|\xi} \\
f_{y}\dfrac{p_{y}}{p_{z}+\big|\big|\textbf{p}\big|\big|\xi}
\end{bmatrix} +
\begin{bmatrix}
c_{x}\\
c_{y}
\end{bmatrix}, \\
\pi^{-1}\big({\bm u}, d\big) &= \dfrac{1}{d}\Bigg(\dfrac{\xi+\sqrt{1+(1-\xi)^2(\hat{u}^2+\hat{v}^2)}}{\hat{u}^2+\hat{v}^2+1} 
\begin{bmatrix}
\hat{u}^2 \\
\hat{v}^2 \\
1
\end{bmatrix}-
\begin{bmatrix}
0 \\
0 \\
\xi
\end{bmatrix} \Bigg),
\end{split}
\end{align*}
\noindent where the subscript~$kf$ refers to keyframe,
\begin{enumerate}
	\item [--] $I_{kf}({\bm u}): \Pi_{kf} \rightarrow \mathbb{R}$~denotes the intensity of any 2D pixel coordinates~${\bm u}$ in the keyframe and $\Pi_{kf} \subset \mathbb{R}^2$ denotes the image domain of keyframe; Detailed explanation regarding keyframe selection and estimation of semi-dense depth maps is given in prior literature~\cite{caruso2015large};
	\item [--] $I_{t}({\bm u}): \Pi_{ns} \rightarrow \mathbb{R}$~denotes the intensity of any 2D pixel coordinates~${\bm u}$ in the current frame and $\Pi_{ns} \subseteq \Pi$ denotes the image domain consisting of non-sky pixels; 
	\item [--] $\pi: \mathbb{R}^3 \rightarrow \Pi_{ns}$ denotes the map from 3D world coordinates, denoted by ${\bm p}=[p_{x},p_{y}, p_{z}]$ to 2D pixel in image frame; 
	\item [--] $w(\Delta{\bm \mu},{\bm u})$ denotes the 3D warp function that unprojects the pixel coordinates~${\bm u}$ and transforms it by a relative state vector $\Delta{\bm \mu}$. The relative state vector~$\Delta{\bm \mu}$ indicates the difference between the current vehicle pose, denoted by ${\bm \mu}_{t}=[\textbf{x},{\bm \psi}]_{t}$ with respect to that of the keyframe, denoted by~${\bm \mu}_{kf}$; Here, $\textbf{x}$~denotes the 3D vehicle position and ${\bm \psi}$~denotes the 3D orientation; ${\textbf R}\in \textrm{SO(3)}$ and ${\textbf t}\in \mathbb{R}^{3}$ denotes the rotation matrix and translation vector of~${\bm \mu}$, respectively; 
	\item [--] $\pi^{-1}:\Pi_{ns} \times \mathbb{R}^{+}\rightarrow \mathbb{R}^{3}$ denotes the inverse mapping of 2D pixel coordinates to 3D world coordinates via an inverse distance represented by $d$. Here, $\hat{u} = (u-c_x)/f_x$ and $\hat{v} = (v-c_y)/f_y$ denotes the transformed 2D pixel coordinates. We calibrate the camera parameters, namely $f_{x}$, $f_{y}$, $c_{x}$, $c_{y}$ and $\xi$ during initialization; 
	\item [--] $d_{kf}({\bm u})$ denotes the inverse distance of the pixel coordinates in the keyframe.
\end{enumerate} 

~\\
\noindent \textit{2. GPS module: }~\\
\indent In the GPS module, considering $N$~visible satellites, we describe the GPS measurement model as 
\begin{align}
\begin{split}
\rho^{k} &=  
\|\textbf{y}^{k}-\textbf{x}\|+\big(c\delta t-c\delta t^{k}\big)+\eta^{k}, \\ 
\end{split}
\label{eq:pseudo}
\end{align}
\noindent where $\textbf{x}$ and $\textbf{y}^{k}$~denotes the 3D position of the vehicle and $k^{th}$~satellite, respectively;~$c\delta t$ and $c\delta t^{k}$ represents the receiver clock bias and $k^{th}$~satellite clock bias corrections, respectively; $\eta^{k}$ represents the measurement noise related to $k^{th}$~satellite. 

We also formulate the measurement covariance of $k^{th}$ satellite via the measured~$C/N_0$ values and the sky area detected via Eq.~\eqref{eq_algo:hybrid} in the vision pre-processing module. Note that the classification of the satellite as either LOS or NLOS depends on the unknown state vector of the vehicle and $k^{th}$ satellite. Therefore, the measurement covariance of $k^{th}$ satellite is given by
\begin{align} \label{eq:cov}
\big(\sigma^{k}({\bm x}_{t}, {\bm y}^{k}_{t})\big)^2= \sqrt{b^{k}+a^{k}\dfrac{1}{(C/N_{0})^{k}}}~\textrm{from~\cite{shytermeja2014proposed}},
\end{align}
\noindent where
\begin{itemize}
	\item[--] ${\bm x}_{t}$ denotes the  vehicle state vector at $t^{th}$~time instant comprising of 3D position, 3D velocity, clock bias, clock drift and 3D attitude, respectively, such that ${\bm x}_{t}=[\textbf{x},~c\delta t,~\dot{\textbf{x}},~c\dot{\delta t},~\bm{\psi}]_{t}$; 
	\item[--] ${\bm y}_{t}^{i}$ denotes the state vector of~$i^{th}$~satellite comprising of its 3D position, 3D velocity, clock bias and clock drift corrections, such that ${\bm y}_{t}^{k}=[\textbf{y}^{k},~c\delta t^{i},~\dot{\textbf{y}}^{k},~c\delta\dot{t}^{k}]_{t},~i\in\{1,\cdots,N\}$;
	\item [--] $b^{k}$ and $a^{k}$ are the vision coefficients, such that $b^{k}=\dfrac{b_{LOS}}{p_{sky}\big(\pi(\textbf{y}^{k})\big)}$ and $a^{k}=\dfrac{a_{LOS}}{p_{sky}\big(\pi(\textbf{y}^{k})\big)}$ when $p_{sky}\big(\pi(\textbf{y}^{k})\big)>\eta$ and $b^{k}=\dfrac{b_{NLOS}}{p_{sky}\big(\pi(\textbf{y}^{k})\big)}$ and $a^{k}=\dfrac{a_{NLOS}}{p_{sky}\big(\pi(\textbf{y}^{k})\big)}$ otherwise; $\eta$ is the pre-defined threshold explained in Eq.~\eqref{eq_algo:hybrid}; $b_{LOS}$, $b_{NLOS}$, $a_{LOS}$ and $a_{NLOS}$ are constant pre-determined coefficients and $\pi(\textbf{y}^{k})$ denotes the projection of the state vector of $k^{th}$ satellite in the image frame.
\end{itemize}

\subsection{Extended graph optimization} \label{sec_algo:graph_opt}
In our extended graph optimization module, our cost function consists of four error terms, namely GPS pseudoranges, non-sky pixel intensities, receiver motion model and satellite orbital model, as follows: 
\begin{align} \label{eq:track_err}
e_{t}(\bm{\theta}_{t}) &=\sum_{k=1}^{N}\Lambda\Big(\big((\bar{r}^{k}_{t}+1)\sigma_{t}^{k}\big)^{-1}\Big|\bm{\rho}_{t}^{k}-h({\bm x}_{t},{\bm y}^{k}_{t})\Big|\Big) + \sum_{k=1}^{N}\Lambda\Big(\big(\hat{\Omega}^{k}_{t}\big)^{-1}\big|\big|{\bm y}_{t}^{i}-f(u^{k}_{t},\bar{\bm y}^{k}_{t-1})\big|\big|\Big)  \\
&~~~+\Lambda\Big(\big(\bar{\chi}_{t}{\bm I}+\hat{\Sigma}_{t}\big)^{-1}\big|\big|{\bm x}_{t}-g(u_{R,t},\bar{\bm x}_{t-1})\big|\big| \Big) + \sum_{{\bm u}\in \Pi_{ns}}\Lambda\Big(\big((\bar{s}_{t-1}({\bm u})+1)\omega_{t}({\bm u})\big)^{-1}\big|\bm{I}_{kf}({\bm u})-\bm{I}_{t}\Big(\pi\big(w(\Delta{\bm \mu}_{t},{\bm u})\big)\Big)\big|\Big), \nonumber 
\end{align}
\noindent where 
\begin{itemize}
	\item[--] $\bm{\theta}_{t}$~denotes the overall state vector comprising of the state vector of the vehicle, GPS satellites and key image pixels in the world frame, given by~$\bm{\theta}_{t}=\{{\bm x}_{t},{\bm y}_{t}^{1},\cdots,{\bm y}_{t}^{N},{\bm p}_{t}^{j}~\forall j\in|\Pi|_{ns}\}$ and is estimated during the graph optimization; 
	\item[--] $\Lambda$~denotes the M-estimator used to transform the corresponding weighted residuals; Details regarding the choice of M-estimator used are explained in our prior work~\cite{bhamidipati2018multiple};
	\item[--] $\bar{r}^{k}_{t-1}$~denotes the fault status associated with the GPS pseudorange of $k^{th}$ satellite and estimated at the past time instant; Similarly, $s_{t}({\bm u})$~denotes the estimated vision fault status of any 2D pixel~${\bm u}\in \Pi_{kf}$ at the previous time instant;
	\item [--] $h$~denotes the GPS measurement model; $g$ denotes the motion model of the receiver and $f$ denotes the satellite orbital model;~$\bar{\bm x}_{t-1}$ and $\bar{\bm y}^{k}_{t-1}$ denotes the estimated state vector of the vehicle and $k^{th}$ satellite, respectively, at the previous time instant;~$u_{R,t}$ and $u^{k}_{t}$~denote the motion control inputs of the vehicle and $k^{th}$ satellite, respectively; 
	\item[--] $\hat{\Sigma}_{t}$ and $\hat{\Omega}^{k}_{t}$~denotes the predicted covariance matrix of the vehicle state vector and $k^{th}$ satellite state vector at the $t^{th}$ time instant; Explanation regarding estimating these covariances is given in our prior work~\cite{bhamidipati2018multiple};
	\item[--] $\sigma_{t}^{k}$ denote the measurement covariance of the $k^{th}$ satellite and is estimated from Eq.~\eqref{eq:cov}; Similarly, $\omega_{t}({\bm u})$~denotes the covariance associated with the intensity of the non-sky pixel~${\bm u}$ and is estimated based on Section~$2.3$ of~\cite{engel2014lsd}.
\end{itemize}

The first three terms in the cost function~$\mathbf{e}_{t}$, given in Eq.~\eqref{eq:track_err}, correspond to the residuals associated with the GPS pseudoranges, satellite ephemeris and vehicle state vector, whose details are provided in our prior work~\cite{bhamidipati2018multiple}. The last term represents the summation of intensity residuals across non-sky pixels based on the vision measurement model explained in Eq.~\eqref{eq_ion:cam_model}. In particular, we perform sub-graph optimization at each instant, as seen in Eq.~\eqref{eq:track_error}, where the cost function is formulated using the past history of measurements.
\begin{align} \label{eq:track_error}
\bm{\bar{\theta}}_{t-T:t} &=\argmin_{\theta_{t-T:t}}\Bigg(\sum_{s=t-T}^{t} e_{s}(\bm{\theta}_{s})\Bigg),
\end{align}
\noindent where $T$~denotes the number of time instants utilized in the sub-graph optimization thread and $\bm{\bar{\theta}}_{t-T:t}$~denotes our SLAM-based IM estimate of the overall state vector computed during the sub-graph optimization. We estimate the key image pixels in the world frame, represented by ${\bm p}_{t}^{j}$, via inverse-mapping defined in Eq.~\eqref{eq_ion:cam_model}. Details regarding mapping that involves periodically executing full-graph optimization is given in our prior work~\cite{bhamidipati2018multiple}. 
 
\subsection{IM for Graph-SLAM framework} \label{sec_ion:fdi_im}
We compute the protection levels associated with the estimated vehicle position using worst-case failure mode slope analysis~\cite{salos2013receiver,joerger2014solution}. This is justified because worst-case failure mode slope is derived for weighted least squares estimator and graph optimization via M-estimator-based Levenberg Marquardt algorithm is also a non-linear weighted least squares problem. However, there are certain design challenges involved in applying worst-case failure slope analysis for the protection level computation of Graph-SLAM framework. Firstly, given that the worst-case failure slope is derived for linear measurement model but the cost function associated with Graph-SLAM is non-linear, we linearize the formulation of graph optimization at the estimated overall state vector. Secondly, Graph-SLAM is a sequential methodology, whereas the worst-case failure slope falls under snapshot technique for integrity monitoring. Therefore, we linearize our graph formulation over not only the current time instant, but over the past time history of measurements so as to incorporate the temporal aspect in protection level computation. Thirdly, the graph optimization for SLAM framework consists of a large number of states and measurements. However, evaluating all possible fault modes associated with the measurements is computationally cumbersome. Therefore, we directly compute a single fault mode based on the measurement fault status estimated via multiple FDI module. 

~\\
\textit{1) Multiple FDI module: } ~\\
\indent Based on the estimated overall state vector from the extended graph optimization explained in Section~\ref{sec_algo:graph_opt}, we independently compute the measurement residuals associated with GPS pseudoranges and non-sky pixel intensities. In our multiple FDI module, we evaluate the GPS residuals by analyzing the temporal correlation of their non-faulty error distribution and vision residuals using spatial correlation across image pixels.  

~\\
\noindent \textbf{GPS faults: }To detect and isolate GPS faults in pseudoranges, we evaluate each residual against an empirical Gaussian distribution, which represents the measurement error distribution during non-faulty conditions. This is justified because we observe that the GPS measurements follow a Gaussian distribution during non-faulty conditions, as explained in our prior work~\cite{bhamidipati2018multiple}. We replicate the non-faulty conditions of GPS measurements by executing the initialization procedure in open-sky conditions. Thereafter, deviation of the measurement residual, denoted by $\Delta\rho^{k}$, from the Cumulative Distribution Function~(CDF) of its empirical Gaussian distribution, denoted by $\Phi_{\Delta\rho}^{k}$, is categorized as a fault and the corresponding fault status~$\bar{r}^{k}_{t}$ is computed in Eq.~\eqref{eq_algo:gps_faults}. The justification regarding the formulation of fault status is explained in our prior work~\cite{bhamidipati2018receivers}. 
\begin{align} \label{eq_algo:gps_faults}
\bar{r}^{k}_{t} = 4\Big(\Phi_{\Delta\rho}^{k}(\Delta \rho^{k})-0.5\Big)^2~~~\forall~ k\in\{1,\cdots,N\}.
\end{align}

\noindent \textbf{Vision faults: }Unlike GPS faults, vision faults caused by data association errors exhibit high spatial correlation across image pixels and low temporal correlation. This is justified because the vision faults are localized to a group of neighboring pixels and are not isolated to a standalone pixel. We developed a superpixel-based piecewise RANSAC technique that performs spatial voting across the image pixels to detect and isolate vision faults. RANSAC~\cite{conte2009vision}, a popular outlier detection method in image processing, estimates the optimal fitting parameters of a model via random sampling of data containing both inliers and outliers.  
\begin{figure}[H]
	\setlength{\belowcaptionskip}{-4pt}
	\centering	\includegraphics[width=0.68\columnwidth]{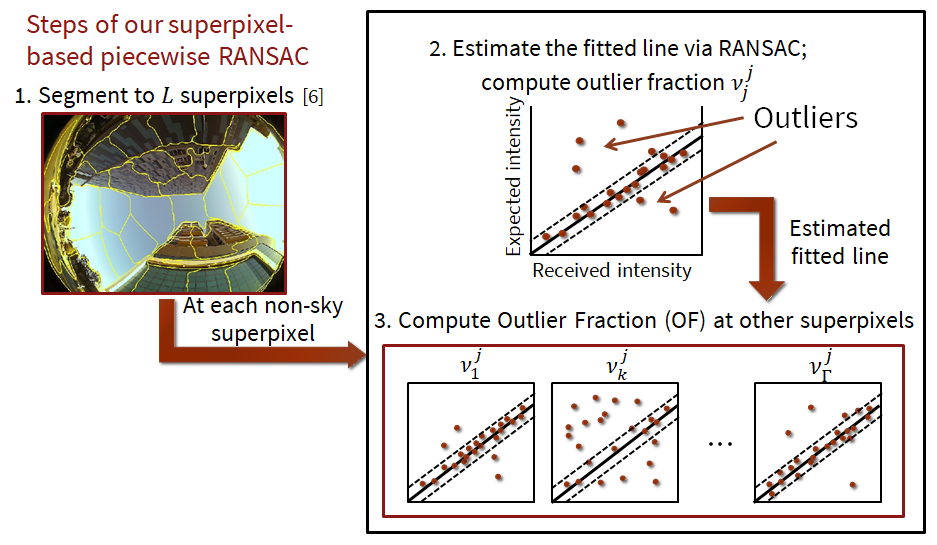}
	\caption{Pipeline for superpixel-based piecewise RANSAC technique used for estimating the vision fault status.}
	\label{fig_algo:vision_faults}
\end{figure}

The steps involved in the superpixel-based piecewise RANSAC technique are described as follows: first, we segment the image into clusters, known as superpixels, based on the color similarity and space proximity between image pixels using superpixel segmentation~\cite{li2015superpixel}. We denote the number of superpixels depicting non-sky pixels to be $\Gamma$, where the total number of superpixels into which the image is segmented is pre-defined during initialization. For each non-sky superpixel, we denote the pixel intensity vector as ${\bm I}^{j}\,\forall j\in\{1,\cdots,\Gamma\}$, which stacks the intensities of pixels within the superpixel. We represent the received intensity, i.e., keyframe pixel intensities Vs expected intensity, i.e., transformed current pixel intensities as a two-Dimensional~(2D) plot. Next, we estimate the fitted line using RANSAC that passes through the optimal set of inliers and thereafter, compute the fraction of outliers in the superpixel, which is represented by $\nu^{j}_{j}$. Then, utilizing the estimated model parameters of the fitted line, we evaluate the corresponding fraction of outliers at all the other non-sky superpixels, denoted by $\nu^{j}_{k}\, \forall k\in\{1,\cdots,\Gamma\}-j$. Finally, the fault status at each superpixel is computed as the product of all the estimated outlier fractions, as seen in Eq.~\eqref{eq_algo:vision_flt}, and the same fault status is assigned to all the pixels within that superpixel. This procedure is repeated for all the non-sky superpixels to compute the fault status of all the non-sky pixels in the keyframe. Our algorithm considers an underlying assumption that there are sufficient number of superpixels to reach a common consensus. If the number of superpixels associated with non-sky pixels is less, such as in open-sky setting, a pre-defined penalty is assigned to the vision fault status. 
\begin{align} \label{eq_algo:vision_flt}
\bar{s}_{t}({\bm u})= \nu^{j}_{1}\cdots \nu^{j}_{\Gamma}~~~\forall {\bm u}\in {\bm I}^{j}
\end{align}

~\\
\textit{2) Protection level computation}~\\
\indent In Eq.~\eqref{eq_ion:overall_meas}, based on the design solutions explained in Section~\ref{sec_ion:fdi_im}.1, we linearize the overall measurement model of the graph optimization framework using first-order approximation. For simplicity, we derive the protection levels using measurements of the current time instant, but the same formulation is applicable for extension to the past history of measurements. 
\begin{align} \label{eq_ion:overall_meas}
\Delta{\bm z} = C\Delta {\bm \theta} + {\bm \eta} + {\bm f},
\end{align}
\noindent where
\begin{enumerate}
	\item [--] $\Delta {\bm z}$~denotes the overall measurement vector that concatenates GPS pseudoranges, control input of vehicle, satellite ephemeris and keyframe pixel intensities against an estimated overall state vector~$\bar{\bm \theta}_{t}$; ${\bm \eta}$ denotes the overall measurement noise; 
	\item [--] $C$~denotes the linearized overall measurement model that vertically stacks the Jacobian associated with GPS pseudoranges, denoted by $H$, vehicle motion model and satellite orbital model, denoted by $A$ and non-sky pixel intensities, denoted by $J$, such that $C = [H,A,J]^{T}$;  
	\item[--] ${\bm f}$~denotes the overall fault vector associated with the overall measurement vector and thereby, stacks measurement faults obtained from individual sensor sources. 
\end{enumerate}

As described in Eq.~\eqref{eq:track_err} of the graph optimization module, we express the M-estimator-based Levenberg Marquardt formulation, which is a weighted non-linear least squares problem, as
\begin{align} \label{eq_ion:overall_state}
\Delta \bar{\bm \theta}_{t} &= K_{t} \Delta{\bm z}_{t}, \nonumber \\
K_{t}&=[V_{t}H_{t}^{T}S_{t}^{-1}, V_{t}A_{t}^{T}R_{t}^{-1}, V_{t}J_{t}^{T}P_{t}^{-1}],
\end{align}
\noindent where 
\begin{enumerate}
	\item [--] $K_{t}$~denotes the estimation matrix of the graph-optimization framework and $V$~denotes the pseudo-inverse matrix, such that $V_{t}=\Big(H_{t}^TS_{t}^{-1}H_{t}+ A_{t}^TR_{t}^{-1}A_{t}+J_{t}^TP_{t}^{-1}J_{t}+\beta\,diag(H_{t}^{T}H_{t}+A_{t}^{T}A_{t}+J_{t}^{T}J_{t})\Big)^{-1}$; 
	\item [--] $S_{t},\,R_{t},\,P_{t}$~denotes the M-estimator-based weight functions for the GPS pseudoranges, vehicle motion model and satellite orbital model, non-sky image pixel intensities, respectively, and evaluated at ${\bm \theta}=\bar{\bm \theta}_{t}$; Details regarding the choice of M-estimator and the corresponding weight functions are explained in our prior work~\cite{bhamidipati2018multiple}; 
	\item [--] $\beta_{n,t}$~denotes the iterative damping factor associated with the Levenberg Marquardt algorithm. 
\end{enumerate}

Next, we define the overall test statistic, denoted by $\zeta$, as the summation of the weighted squared residuals across all the measurements. We consider an assumption that the overall test statistic is chi-square distributed, denoted by $\chi^{2}_{k}$ under non-faulty conditions and non-central chi-squared, denoted by $\chi^{2}_{k,\lambda}$, under the presence of GPS faults or vision faults or both.  
\begin{align} \label{eq_algo:test_stat}
\zeta &= \big(\Delta {\bm z}-C\Delta \bar{\bm \theta}\big)^{T} \big(\Delta {\bm z}-C\Delta \bar{\bm \theta}\big),
\end{align}
\noindent such that
\begin{align}
\zeta = \begin{cases}
\chi^{2}_{k} & {\bm f}={\bm 0}~\textrm{or non-faulty},\\
\chi^{2}_{k,\lambda} & \textrm{otherwise}. 
\end{cases}
\end{align}
\noindent where $k$~denotes the number of redundant measurements, i.e., difference between the number of overall measurements, denoted by $n$ and overall states, denoted by $l$, such that $k=n-l$. $\lambda$~indicates the non-centrality parameter associated with the overall test statistic during faulty conditions.

\begin{figure}[H]
	\setlength{\belowcaptionskip}{-4pt}
	\centering	\includegraphics[width=0.4\columnwidth]{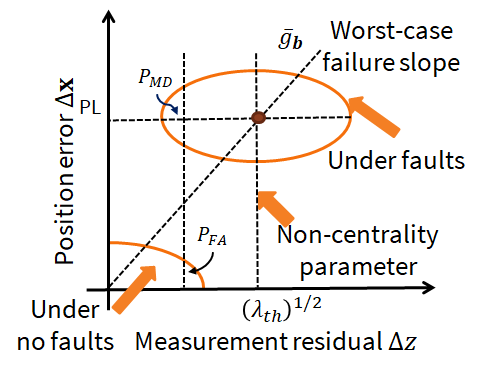}
	\caption{Protection levels computed as the intersection of worst-case failure mode slope and non-centrality parameter~\cite{salos2013receiver}.}
	\label{fig_algo:prot_levels}
\end{figure}

According to the worst-case failure mode slope analysis~\cite{salos2013receiver}, as seen in Fig.~\ref{fig_algo:prot_levels}, the protection level is calculated as the projection in the position domain of the measurement faults that would generate a non-centrality parameter~$\lambda=\lambda_{th}$ in the overall test statistic~$\zeta$ with the maximum slope. In particular, the non-centrality parameter~$\lambda_{th}$ is estimated from the false-alarm, denoted by~$p_{FA}$ and mis-detection rates, denoted by~$p_{MD}$, which are set according to the pre-defined integrity requirements. 

In Eq.~\eqref{eq_algo:fault_mode}, we formulate the measurement fault mode, denoted by ${\bm b}_{t}$, using GPS and vision fault status estimated in Eqs.~\eqref{eq_algo:gps_faults} and~\eqref{eq_algo:vision_flt}. For this, we consider a pre-defined fault threshold, denoted by $\kappa$, such that if the fault status is above $\kappa$, the measurement is flagged as faulty in the computation of protection levels. Given that we consider measurement faults in only GPS and vision, the fault entries of receiver and satellite motion models are set to zero for this work. However, the corresponding fault vector, which comprises of the exact measurement fault magnitudes, is still unknown. According to~\cite{salos2013receiver}, for a given fault mode, the worst case fault direction that maximizes the integrity risk, is the one that maximizes the failure mode slope, which is seen in Fig.~\ref{fig_algo:prot_levels} and denoted by~$g_{\bm b}$. In this context, we define the square of failure mode slope, denoted by~$g^{2}_{\bm b}$, as the ratio of squared state estimation error in position of the vehicle over the overall test statistic. Using the linearized equations seen in Eqs.~\eqref{eq_ion:overall_meas},~\eqref{eq_ion:overall_state} and~\eqref{eq_algo:test_stat}, we derive the failure slope for the graph optimization framework in terms of unknown fault vector. For this, we consider $CKC\approx \textbf{I}$, which is valid approximation after the iterative convergence of the graph optimization at any time instant since $\beta<<0$. 
\begin{align} \label{eq_algo:fault_mode}
{\bm b}_{t} = \bigg[\mathds{1}_{\big\{\bar{r}^{1}_{t}>\kappa\big\}},\cdots,\mathds{1}_{\big\{\bar{r}^{1}_{t}>\kappa\big\}},{\bm 0},\mathds{1}_{\big\{\bar{s}({\bm u})_{t}>\kappa\big\}}~\forall {\bm u}\in \Pi_{ns} \bigg]_{t}
\end{align} 

Considering $n_{\bm b}$ to be the number of non-zero entries in the fault mode~${\bm b}$ estimated via multiple FDI module, we define fault matrix, denoted by $B_{\bm b}$, as $B_{\bm b}=[{\bm I}_{n_{\bm b}},{\bm 0}_{n-n_{\bm b}}]^{T}$ and next, re-arrange the rows of the $m_{\epsilon}$ and $M_{\zeta}$ matrices to match the rows of the fault matrix. Thereafter, we define a transformed fault vector, denoted by~${\bm f}_{\zeta}$, such that ${\bm f}=B_{\bm b}M_{\zeta}{\bm f}_{\zeta}$. Based on the above-mentioned steps, we describe the failure slope formulation of Graph-SLAM framework in Eq.~\eqref{eq_algo:fail_slp}.

\begin{align} \label{eq_algo:fail_slp}
g^{2}_{\bm b} &= \dfrac{{\bm \epsilon}^{T}{\bm \epsilon}}{\zeta} = \dfrac{\big(\Delta {\bm \theta}-\Delta \bar{\bm \theta}\big)^{T}\big(\Delta {\bm \theta}-\Delta \bar{\bm \theta}\big)}{\big(\Delta {\bm z}-C\Delta \bar{\bm \theta}\big)^{T} \big(\Delta {\bm z}-C\Delta \bar{\bm \theta}\big)},\nonumber \\
&= \dfrac{{\bm f}^{T}\Big[\big(\alpha^TK\big)^{T}\big(\alpha^TK\big)\Big]{\bm f}}{{\bm f}^{T}\Big[\big(\textbf{I}-CK\big)^{T}\big(\textbf{I}-CK\big)\Big] {\bm f}}, \\
&= \dfrac{{\bm f}_{\zeta}^{T}M^{T}_{\zeta}m_{\bm \epsilon}m_{\bm \epsilon}^{T}M_{\zeta}{\bm f}_{\zeta}}{{\bm f}_{\zeta}^{T}{\bm f}_{\zeta}}.\nonumber  
\end{align}
\noindent where $\alpha$ extracts the vehicle 3D position from the overall state vector~${\bm \theta}$, such that $\alpha^{T}=[\bm{1}_{3\times 1},\bm{0}_{(l-3)\times 1}]$, $M_{\zeta}$~denotes the residual matrix, such that $M_{\zeta}=\Big(B_{\bm b}^{T}\Big[\big(\textbf{I}-CK\big)^{T}\big(\textbf{I}-CK\big)\Big]B_{\bm b}\Big)^{-1/2}$ and $m_{\bm \epsilon}$~represents the state gain matrix, such that $m_{\bm \epsilon}=B^{T}_{\bm b}\alpha^TK$.

Referring to~\cite{joerger2014solution}, for a given fault mode but unknown fault vector, the worst-case failure slope equals the maximum eigenvalue of the corresponding failure slope formulation. Therefore, we express the worst-case failure slope of the Graph-SLAM framework as
\begin{align}
\bar{g}^2_{\bm b} = m_{\bm \epsilon}^{T}M_{\zeta}M_{\zeta}^{T}m_{\bm \epsilon}.
\end{align}

Next, we compute protection level~$\bar{\chi}_{t}$, seen in Eq.~\eqref{eq_algo:pl} as the y-coordinate that corresponds to the integrity metric~$\lambda_{th}$ along the line passing through the origin and with slope given by~$\bar{g}_{\bm b}^{2}$.
\begin{align} \label{eq_algo:pl}
\bar{\chi}_{t} = \sqrt{\lambda_{th}\bar{g}_{\bm b}^{2}}
\end{align}
\begin{figure}[H]
	\setlength{\belowcaptionskip}{-4pt}
	\centering	\includegraphics[width=0.6\columnwidth]{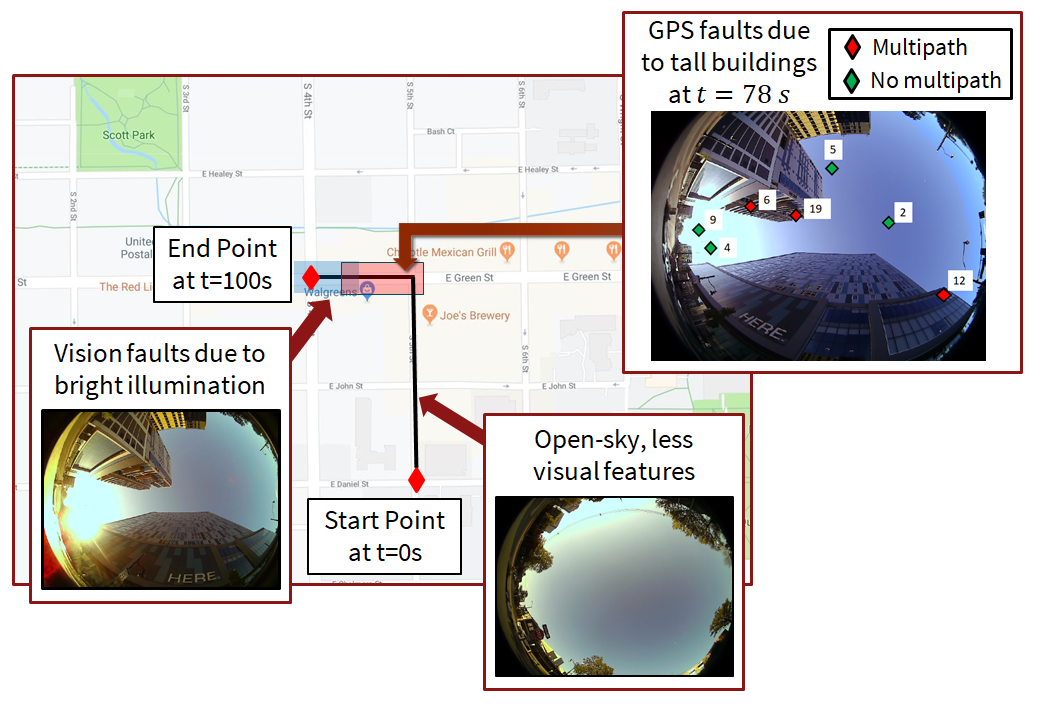}
	\caption{Route taken by a ground vehicle during the experiment conducted for $100~$s. Between $t=70-100~$s, the vehicle experiences GPS faults due to multipath and vision faults due to illumination variations. At $t=78~$s, the overlap of the skyplot of GPS satellites with the fish-eye image shows the multipath affected GPS measurements. }
	\label{fig_exp:route}
\end{figure}

\section{Experiment Results}
We validate the performance of the proposed SLAM-based IM algorithm that utilizes both GPS and fish-eye camera. We conduct real-world experiments on a moving ground vehicle in the semi-urban area of Champaign, IL along the route shown in Fig.~\ref{fig_exp:route}. Our experimental setup comprises of a commercial off-the-shelf GPS receiver and a fish-eye camera fitted with $180^{\circ}$ FOV lens. During $t=0-70~$s, the ground vehicle operates in open-sky conditions, thereby experiencing no GPS faults but less visual features. In Fig.~\ref{fig_exp:route}, the blue highlighted region suffers from vision challenges, namely illumination variations due to sunlight and shadows, that causes data association errors across images. Similarly, the red highlighted region is enclosed with tall buildings that leads to multipath effects in the GPS measurements. For instance, at $t=78~$s we showcase the true overlap of the GPS satellite positions over the fish-eye image, where $3$ out of the $7$ visible satellites are affected by multipath.  

\begin{figure}[H]
	\setlength{\belowcaptionskip}{-4pt}
	\centering	\includegraphics[width=0.6\columnwidth]{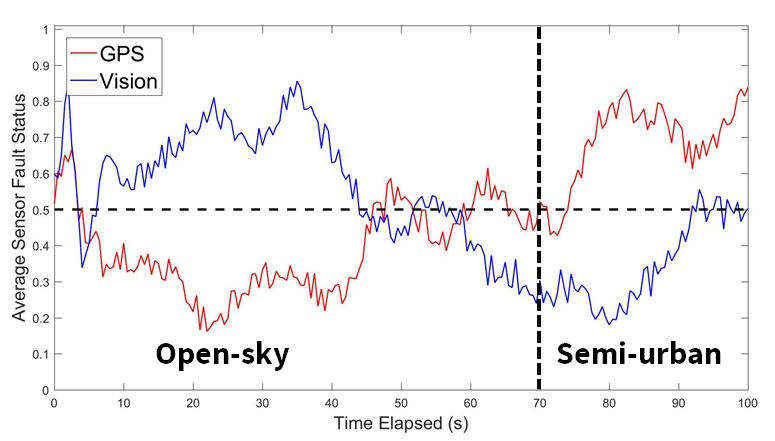}
	\caption{Performance of our multiple FDI module via average fault status of GPS pseudoranges, indicated in red and vision superpixels, indicated in blue. When the ground vehicle navigates through the semi-urban region, i.e., for $t>70~$s, the average fault status associated with GPS is high due to multipath, whereas vision is low due to rich features.}
	\label{fig_exp:avg_fault_prob}
\end{figure}

Fig.~\ref{fig_exp:avg_fault_prob} shows the average fault status of GPS pseudoranges and vision superpixels, as indicated in red and blue, respectively. Given that the ground vehicle navigates in open-sky conditions for $t<70~$s, the average GPS fault status estimated via our multiple FDI module is low, whereas the average vision fault status is high due to the feature-less surroundings. As the vehicle passes through the red highlighted region shown in Fig.~\ref{fig_exp:route} that represents the semi-urban area, the average fault status of vision is low but that of GPS increases due to multipath. 

\begin{figure}[h]
	\centering
	\begin{subfigure}[b]{0.355\textwidth}
		\includegraphics[width=\textwidth]{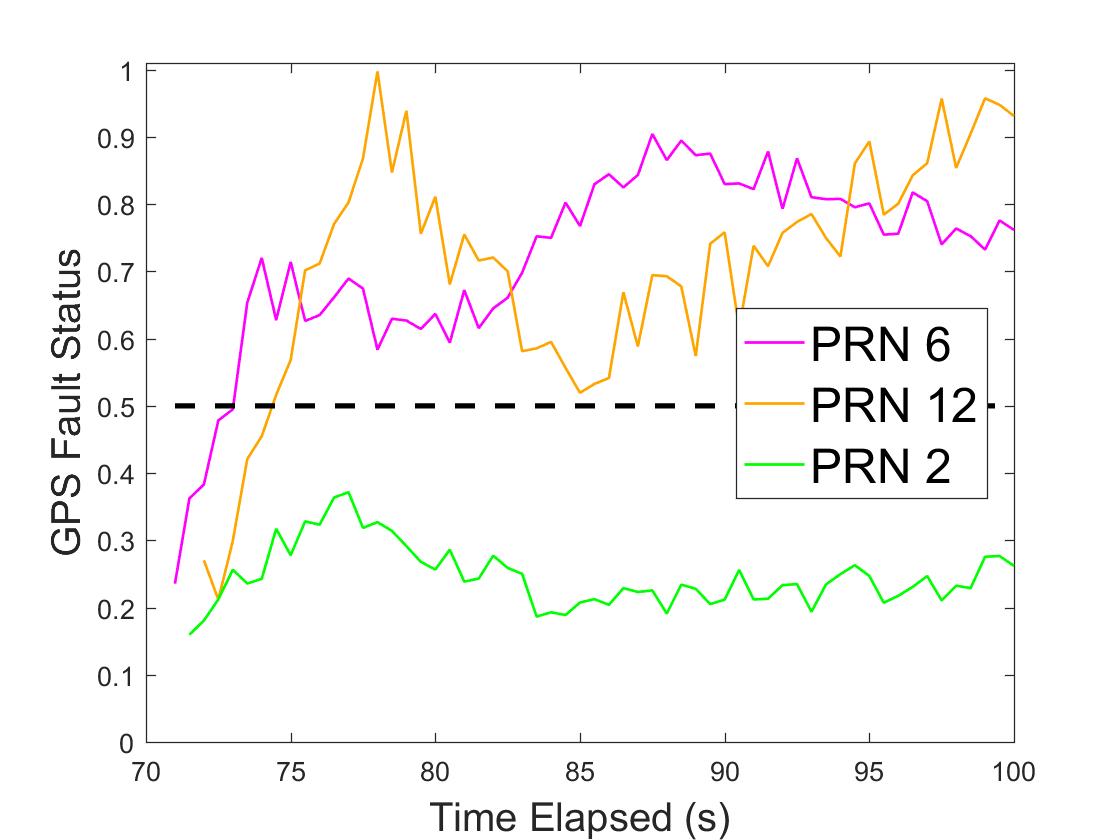}
		\caption{GPS fault status of PRN $6,\,12,\,2$}
		\label{fig_exp:gps_faults}
	\end{subfigure}
	\begin{subfigure}[b]{0.355\textwidth}
		\includegraphics[width=\textwidth]{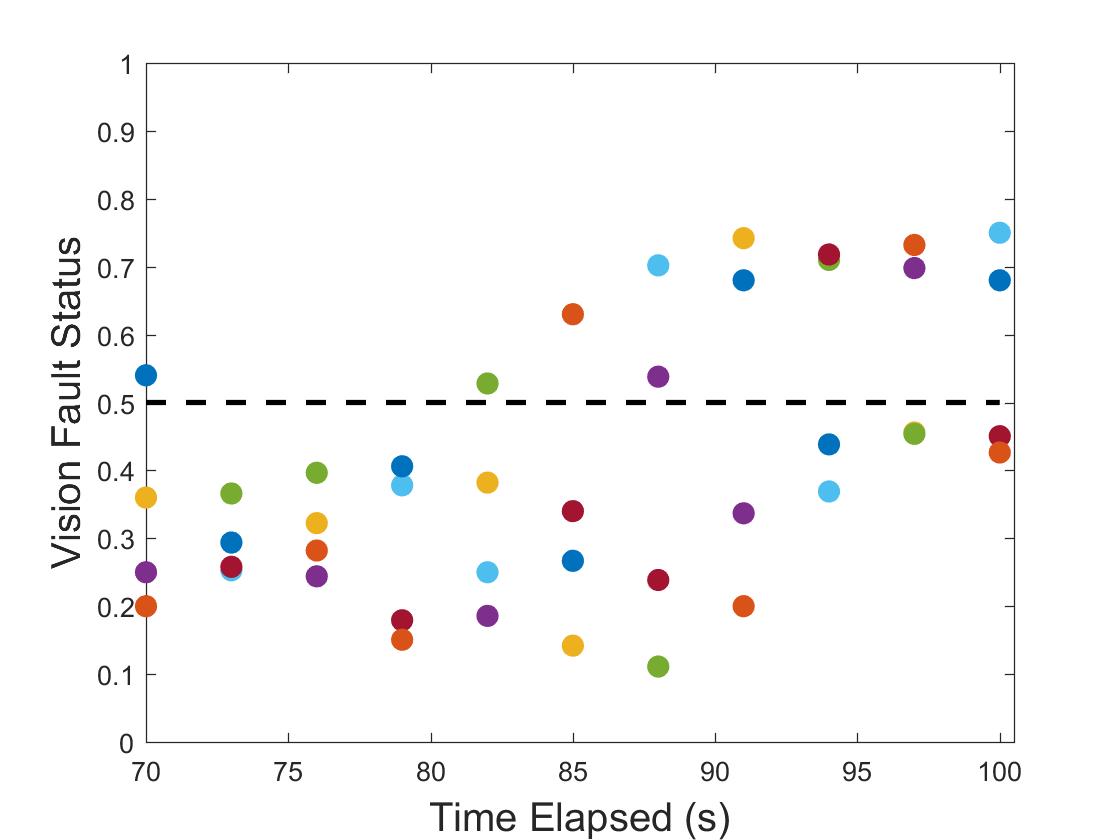}
		\caption{Vision fault status of superpixels}
		\label{fig_exp:vision_faults}
	\end{subfigure}
	\caption{Estimated fault status of (a) GPS measurements and (b) vision superpixels during $t=70-100~$s, i.e., when the ground vehicle navigates through the semi-urban area. In (a), our multiple FDI module successfully detects satellites with PRN $6,\,12$ as faulty while accurately estimating the PRN~$2$ as non-faulty. In (b), where each marker indicates a superpixel, the trend of fault status associated with superpixels is low given the rich features but later, increases due to illumination changes. }
\end{figure}

We further analyze the performance of our multiple FDI module in the challenging semi-urban area, i.e., for $t>70~$s during which the ground vehicle experiences GPS faults due to multipath and vision faults due to illumination variations. Fig.~\ref{fig_exp:gps_faults} plots that the individual GPS fault status of $3$ out of the $7$ visible satellites with PRNs $6,\,12$, and $2$. In accordance with the skyplot shown in Fig.~\ref{fig_exp:route}, our proposed SLAM-based IM algorithm successfully flags the satellites with PRN $6$ and $12$ as faulty while accurately estimating the high-elevation satellite with PRN~$2$ as non-faulty. During the same duration, we also analyze the vision fault status associated with the superpixels. In Fig.~\ref{fig_exp:vision_faults}, at each time instant, we plot the top four fault status of the superpixels, such that each marker represents a superpixel. We observe that in urban region, the value of the associated vision fault status decreases due to feature-rich tall buildings in urban areas. However, when the vehicle enters the blue highlighted region seen in Fig.~\ref{fig_exp:vision_faults}, the illumination variations induced by the bright sunlight causes the fault status associated with certain superpixels to shown an increasing trend. 

\begin{figure}[h]
	\centering
	\begin{subfigure}[b]{0.425\textwidth}
		\includegraphics[width=\textwidth]{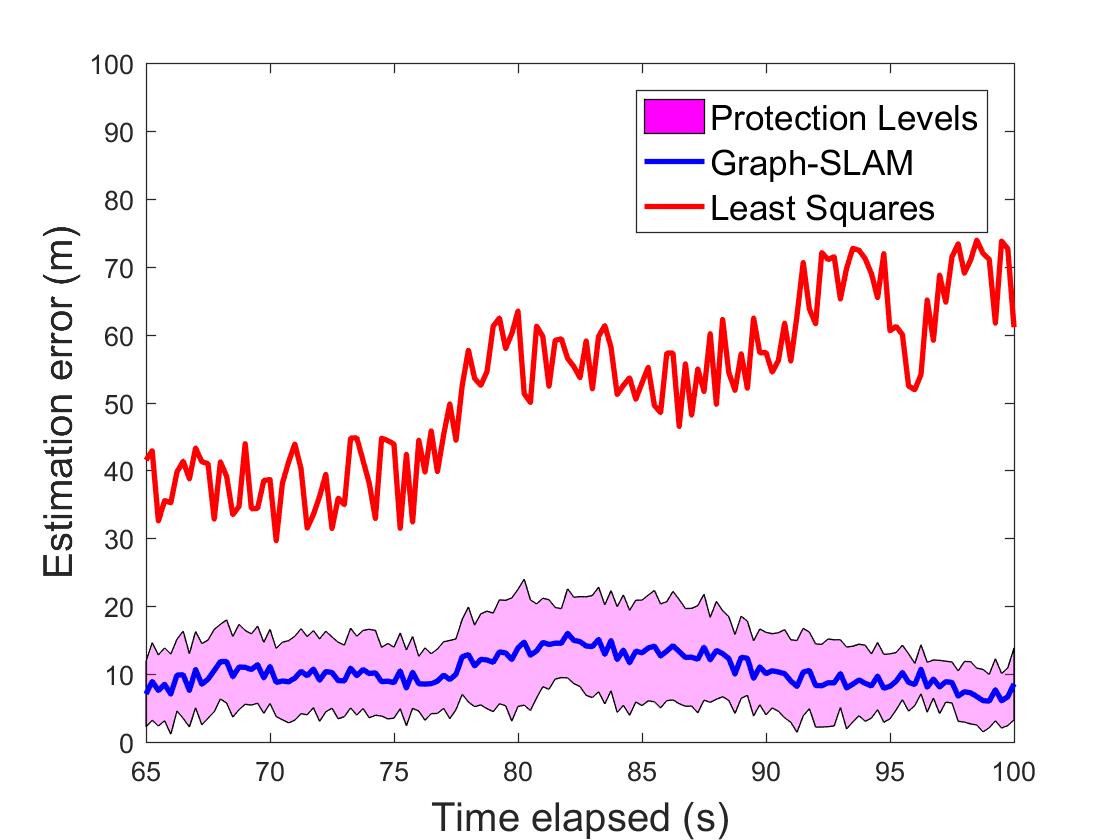}
		\caption{SLAM-based IM via GPS and fish-eye camera}
		\label{fig_exp:PL_multisensor}
	\end{subfigure}
	\begin{subfigure}[b]{0.425\textwidth}
		\includegraphics[width=\textwidth]{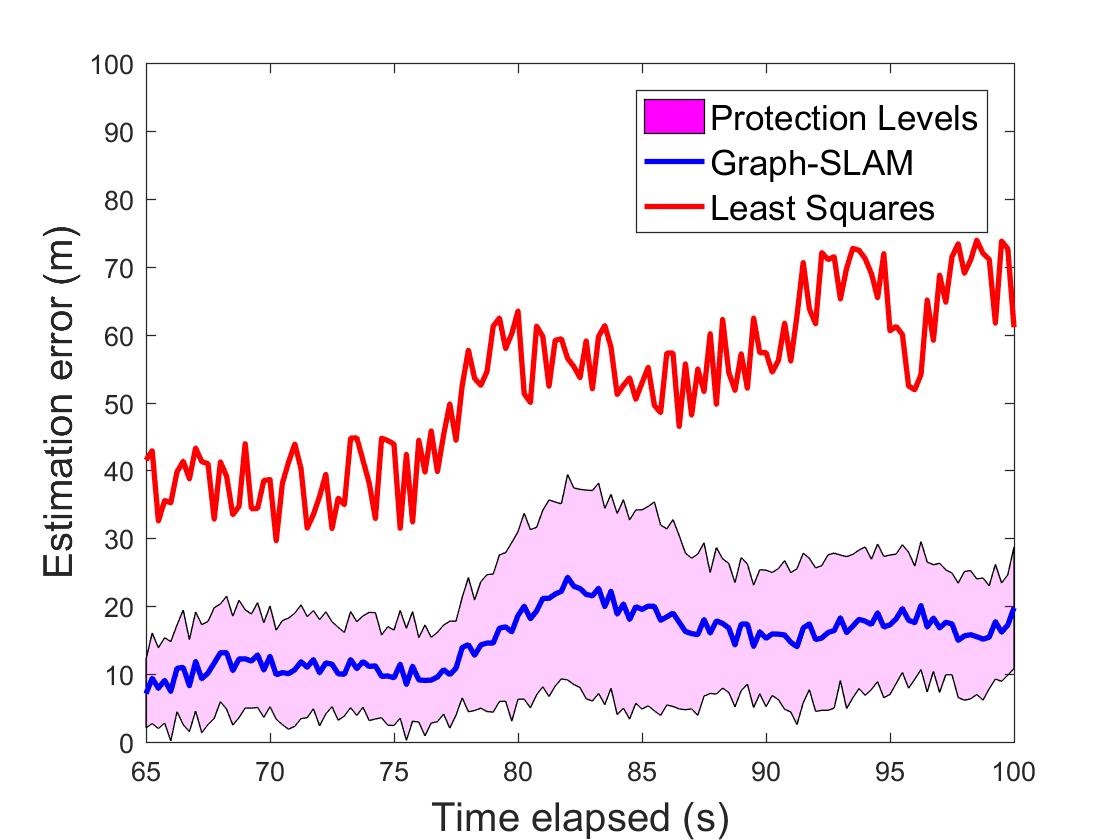}
		\caption{SLAM-based IM via GPS-only}
		\label{fig_exp:PL_GPSonly}
	\end{subfigure}
	\caption{Comparison of SLAM-based IM: (a)~using GPS and fish-eye camera; (b)~using GPS-only. Lower localization errors and tighter protection levels are achieved via GPS and fish-eye camera as compared to GPS-only.}
\end{figure}

We demonstrate the improved performance of the SLAM-based IM algorithm that utilizes GPS and fish-eye camera seen in Fig.~\ref{fig_exp:PL_multisensor}, as compared to the SLAM-based IM algorithm that utilizes GPS-only seen in Fig.~\ref{fig_exp:PL_GPSonly}. By utilizing GPS and fish-eye camera, we demonstrate higher localization accuracy, with an Root Mean Squared Error~(RMSE) of $8.8~$m and standard deviation of $1.73~$m, as compared to employing GPS-only that shows an RMSE of $16.2~$m and standard deviation of $2.86~$m. We also validate that the lower mean size of protection levels are estimated using GPS and fish-eye camera, i.e. $6.5~$m than using GPS-only, i.e., $10.5~$m thereby, achieving tighter protection levels.     

\section{Conclusions}
We proposed a Simultaneous Localization and Mapping~(SLAM)-based Integrity Monitoring~(IM) algorithm using GPS and fish-eye camera that estimates the protection levels of the Graph-SLAM framework while accounting for multiple faults in GPS and vision. We developed hybrid sky detection algorithm to distinguish the non-sky and sky pixels, which are later used in graph optimization and GPS measurement covariance, respectively. By utilizing the GPS pseudoranges, non-sky pixel intensities, receiver and satellite motion model, we performed graph optimization via M-estimator-based Levenberg Marquardt algorithm. We simultaneously estimated the state vector of the vehicle, GPS satellites and key image pixels in the world frame. We estimated the fault mode vector by independently evaluating the measurement residuals against an empirical Gaussian distribution for GPS faults and using our developed superpixel-based piecewise RANSAC for vision faults. We computed the protection levels via the worst-case failure slope analysis that estimates the maximum eigenvalue associated with the failure slope formulation of the linearized Graph-SLAM framework.  

We conducted real-world experiments using a ground vehicle in a semi-urban region to analyze the performance our proposed SLAM-based IM algorithm that utilizes GPS and fish-eye camera. We successfully detected and isolated multiple measurement faults in GPS and vision. We demonstrated higher localization accuracy using our proposed algorithm with an RMSE of $8.8~$m and standard deviation of $1.73~$m, as compared to GPS-only that shows an RMSE of $16.2~$m and standard deviation of $2.86~$m. We also validated that the mean size of protection levels estimated using GPS and fish-eye camera, i.e. $6.5~$m is lower than using GPS-only, i.e., $10.5~$m.    

\bibliographystyle{ieeetran}
\bibliography{IEEEabrv,mybiblibrary}

\begin{thebibliography}{10}
\providecommand{\url}[1]{#1}
\csname url@rmstyle\endcsname
\providecommand{\newblock}{\relax}
\providecommand{\bibinfo}[2]{#2}
\providecommand\BIBentrySTDinterwordspacing{\spaceskip=0pt\relax}
\providecommand\BIBentryALTinterwordstretchfactor{4}
\providecommand\BIBentryALTinterwordspacing{\spaceskip=\fontdimen2\font plus
\BIBentryALTinterwordstretchfactor\fontdimen3\font minus
  \fontdimen4\font\relax}
\providecommand\BIBforeignlanguage[2]{{%
\expandafter\ifx\csname l@#1\endcsname\relax
\typeout{** WARNING: IEEEtran.bst: No hyphenation pattern has been}%
\typeout{** loaded for the language `#1'. Using the pattern for}%
\typeout{** the default language instead.}%
\else
\language=\csname l@#1\endcsname
\fi
#2}}

\bibitem{ochieng2003gps}
W.~Y. Ochieng, K.~Sauer, D.~Walsh, G.~Brodin, S.~Griffin, and M.~Denney, ``Gps
  integrity and potential impact on aviation safety,'' \emph{The journal of
  navigation}, vol.~56, no.~1, pp. 51--65, 2003.

\bibitem{joerger2017towards}
M.~Joerger and M.~Spenko, ``Towards navigation safety for autonomous cars,''
  \emph{Inside GNSS}, 2017.

\bibitem{hol2011sensor}
J.~D. Hol, ``Sensor fusion and calibration of inertial sensors, vision,
  ultra-wideband and gps,'' Ph.D. dissertation, Link{\"o}ping University
  Electronic Press, 2011.

\bibitem{krishnaswamy2008sensor}
K.~Krishnaswamy, ``Sensor fusion for navigation,'' Aug.~14 2008, uS Patent App.
  11/673,906.

\bibitem{miro2006towards}
J.~V. Miro, W.~Zhou, and G.~Dissanayake, ``Towards vision based navigation in
  large indoor environments,'' in \emph{2006 IEEE/RSJ International Conference
  on Intelligent Robots and Systems}.\hskip 1em plus 0.5em minus 0.4em\relax
  IEEE, 2006, pp. 2096--2102.

\bibitem{velaga2012map}
N.~R. Velaga, M.~A. Quddus, A.~L. Bristow, and Y.~Zheng, ``Map-aided integrity
  monitoring of a land vehicle navigation system,'' \emph{IEEE Transactions on
  Intelligent Transportation Systems}, vol.~13, no.~2, pp. 848--858, 2012.

\bibitem{walter2008shaping}
T.~Walter, J.~Blanch, P.~Enge, B.~Pervan, and L.~Gratton, ``Shaping aviation
  integrity: Two raims for safety,'' \emph{GPS World}, vol.~19, no.~4, 2008.

\bibitem{binjammaz2013gps}
T.~Binjammaz, A.~Al-Bayatti, and A.~Al-Hargan, ``Gps integrity monitoring for
  an intelligent transport system,'' in \emph{2013 10th Workshop on
  Positioning, Navigation and Communication (WPNC)}.\hskip 1em plus 0.5em minus
  0.4em\relax IEEE, 2013, pp. 1--6.

\bibitem{li2017lane}
F.~Li, P.~Bonnifait, J.~Ibanez-Guzman, and C.~Zinoune, ``Lane-level
  map-matching with integrity on high-definition maps,'' in \emph{2017 IEEE
  Intelligent Vehicles Symposium (IV)}.\hskip 1em plus 0.5em minus 0.4em\relax
  IEEE, 2017, pp. 1176--1181.

\bibitem{toledo2009lane}
R.~Toledo-Moreo, D.~B{\'e}taille, and F.~Peyret, ``Lane-level integrity
  provision for navigation and map matching with gnss, dead reckoning, and
  enhanced maps,'' \emph{IEEE Transactions on Intelligent Transportation
  Systems}, vol.~11, no.~1, pp. 100--112, 2009.

\bibitem{bhamidipati2018multiple}
S.~Bhamidipati and G.~X. Gao, ``Multiple gps fault detection and isolation
  using a graph-slam framework,'' in \emph{31st International Technical Meeting
  of the Satellite Division of the Institute of Navigation, ION GNSS+
  2018}.\hskip 1em plus 0.5em minus 0.4em\relax Institute of Navigation, 2018,
  pp. 2672--2681.

\bibitem{bhamidipati2018receivers}
------, ``Distributed cooperative slam-based integrity monitoring via a network
  of receivers,'' in \emph{32st International Technical Meeting of the
  Satellite Division of the Institute of Navigation, ION GNSS+ 2019}.\hskip 1em
  plus 0.5em minus 0.4em\relax Institute of Navigation, 2019 (Accepted).

\bibitem{cadena2016past}
C.~Cadena, L.~Carlone, H.~Carrillo, Y.~Latif, D.~Scaramuzza, J.~Neira, I.~Reid,
  and J.~J. Leonard, ``Past, present, and future of simultaneous localization
  and mapping: Toward the robust-perception age,'' \emph{IEEE Transactions on
  robotics}, vol.~32, no.~6, pp. 1309--1332, 2016.

\bibitem{latif2014robust}
Y.~Latif, C.~Cadena, and J.~Neira, ``Robust graph slam back-ends: A comparative
  analysis,'' in \emph{2014 IEEE/RSJ International Conference on Intelligent
  Robots and Systems}.\hskip 1em plus 0.5em minus 0.4em\relax IEEE, 2014, pp.
  2683--2690.

\bibitem{salos2013receiver}
D.~Sal{\'o}s, A.~Martineau, C.~Macabiau, B.~Bonhoure, and D.~Kubrak, ``Receiver
  autonomous integrity monitoring of gnss signals for electronic toll
  collection,'' \emph{IEEE transactions on intelligent transportation systems},
  vol.~15, no.~1, pp. 94--103, 2013.

\bibitem{joerger2014solution}
M.~Joerger, F.-C. Chan, and B.~Pervan, ``Solution separation versus
  residual-based raim,'' \emph{NAVIGATION: Journal of the Institute of
  Navigation}, vol.~61, no.~4, pp. 273--291, 2014.

\bibitem{shytermeja2014proposed}
E.~Shytermeja, A.~Garcia-Pena, and O.~Julien, ``Proposed architecture for
  integrity monitoring of a gnss/mems system with a fisheye camera in urban
  environment,'' in \emph{International Conference on Localization and GNSS
  2014 (ICL-GNSS 2014)}.\hskip 1em plus 0.5em minus 0.4em\relax IEEE, 2014, pp.
  1--6.

\bibitem{lashley2010valid}
M.~Lashley, D.~M. Bevly, and J.~Y. Hung, ``A valid comparison of vector and
  scalar tracking loops,'' in \emph{IEEE/ION Position, Location and Navigation
  Symposium}.\hskip 1em plus 0.5em minus 0.4em\relax IEEE, 2010, pp. 464--474.

\bibitem{shevlyakov2008redescending}
G.~Shevlyakov, S.~Morgenthaler, and A.~Shurygin, ``Redescending m-estimators,''
  \emph{Journal of Statistical Planning and Inference}, vol. 138, no.~10, pp.
  2906--2917, 2008.

\bibitem{lourakis2005brief}
M.~I. Lourakis \emph{et~al.}, ``A brief description of the levenberg-marquardt
  algorithm implemented by levmar,'' \emph{Foundation of Research and
  Technology}, vol.~4, no.~1, pp. 1--6, 2005.

\bibitem{li2015superpixel}
Z.~Li and J.~Chen, ``Superpixel segmentation using linear spectral
  clustering,'' in \emph{Proceedings of the IEEE Conference on Computer Vision
  and Pattern Recognition}, 2015, pp. 1356--1363.

\bibitem{conte2009vision}
G.~Conte and P.~Doherty, ``Vision-based unmanned aerial vehicle navigation
  using geo-referenced information,'' \emph{EURASIP Journal on Advances in
  Signal Processing}, vol. 2009, p.~10, 2009.

\bibitem{wang2010new}
C.-y. Wang, L.-l. Li, F.-p. Yang, and H.~Gong, ``A new kind of adaptive
  weighted median filter algorithm,'' in \emph{2010 International Conference on
  Computer Application and System Modeling (ICCASM 2010)}, vol.~11.\hskip 1em
  plus 0.5em minus 0.4em\relax IEEE, 2010, pp. V11--667.

\bibitem{gao2010improved}
W.~Gao, X.~Zhang, L.~Yang, and H.~Liu, ``An improved sobel edge detection,'' in
  \emph{2010 3rd International Conference on Computer Science and Information
  Technology}, vol.~5.\hskip 1em plus 0.5em minus 0.4em\relax IEEE, 2010, pp.
  67--71.

\bibitem{haque2008hybrid}
M.~Haque, M.~Murshed, and M.~Paul, ``A hybrid object detection technique from
  dynamic background using gaussian mixture models,'' in \emph{2008 IEEE 10th
  Workshop on Multimedia Signal Processing}.\hskip 1em plus 0.5em minus
  0.4em\relax IEEE, 2008, pp. 915--920.

\bibitem{moghaddam2012adotsu}
R.~F. Moghaddam and M.~Cheriet, ``Adotsu: An adaptive and parameterless
  generalization of otsu's method for document image binarization,''
  \emph{Pattern Recognition}, vol.~45, no.~6, pp. 2419--2431, 2012.

\bibitem{caruso2015large}
D.~Caruso, J.~Engel, and D.~Cremers, ``Large-scale direct slam for
  omnidirectional cameras,'' in \emph{2015 IEEE/RSJ International Conference on
  Intelligent Robots and Systems (IROS)}.\hskip 1em plus 0.5em minus
  0.4em\relax IEEE, 2015, pp. 141--148.

\bibitem{engel2014lsd}
J.~Engel, T.~Sch{\"o}ps, and D.~Cremers, ``Lsd-slam: Large-scale direct
  monocular slam,'' in \emph{European conference on computer vision}.\hskip 1em
  plus 0.5em minus 0.4em\relax Springer, 2014, pp. 834--849.

\end{thebibliography}

\end{document}